\documentclass[10pt,twocolumn,letterpaper]{article}

\usepackage{cvpr}              

\usepackage[accsupp]{axessibility}  

\definecolor{cvprblue}{rgb}{0.21,0.49,0.74}
\usepackage[pagebackref,breaklinks,colorlinks,allcolors=cvprblue]{hyperref}

\usepackage{graphicx}
\graphicspath{{FIGURES/}}

\def\paperID{9076} 
\def\confName{CVPR}
\def\confYear{2026}

\title{Mirror Illusion Art}

\author{
  Xiaopei Zhu\textsuperscript{1 *} \quad
  Zeyuan Li\textsuperscript{2 *} \quad
  Jun Zhu\textsuperscript{1,3} \quad
  Xiaolin Hu\textsuperscript{1,3,4 †} \\
  \textsuperscript{1}Department of Computer Science and Technology, BNRist, Tsinghua University \\
  \textsuperscript{2}Huazhong University of Science and Technology \\
  \textsuperscript{3}IDG/McGovern Institute for Brain Research, Tsinghua University \\
  \textsuperscript{4}Chinese Institute for Brain Research (CIBR)
}

\begin{document}

\twocolumn[{%
\maketitle
\vspace{-5mm}
\begin{center}
  \includegraphics[width=\linewidth,height=0.32\textheight,keepaspectratio]{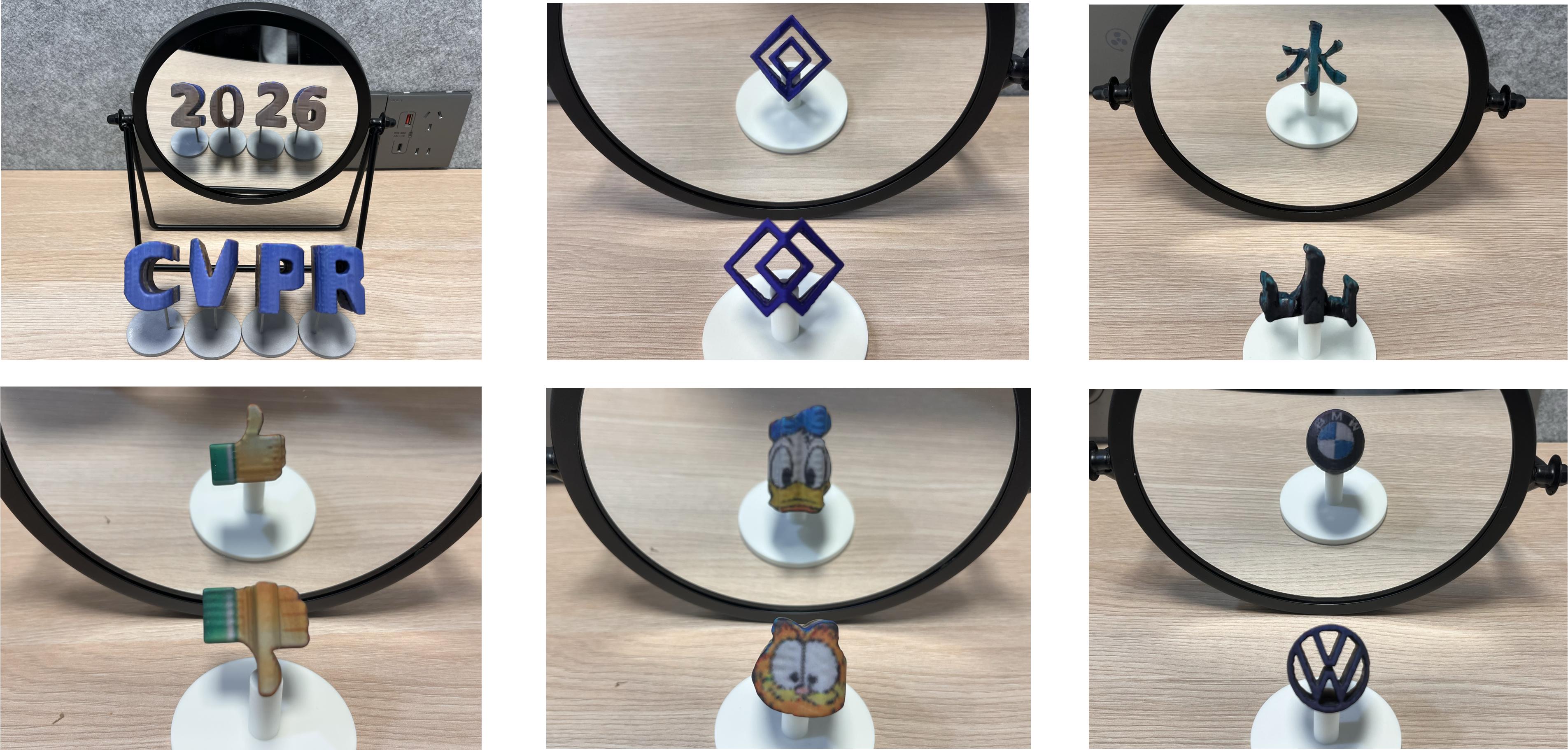}
  \captionof{figure}{Physical visualization of Mirror Illusion Arts designed by our method. Interestingly, it seems that the front view of the 3D object and its mirror view ``do not appear'' to belong to the same object. See \textit{Supplementary Video} for a full-angle visualization of these objects.}
  \label{fig1}
\end{center}
}]
\begingroup
\renewcommand\thefootnote{\fnsymbol{footnote}}
\footnotetext[1]{\footnotesize Equal contribution.}
\footnotetext[2]{\footnotesize Corresponding author.}
\endgroup

\begin{abstract}
Mirror Illusion Art is a novel reflection-conditioned 3D illusion where one object yields two target appearances (front and mirror). The task is formulated as inverse design from two target 2D images (front and mirror) to a printable 3D object with geometry and texture. Prior topology-driven and shadow-based approaches demand substantial manual effort, optimize shape only, and often yield non-smooth or incomplete geometry. To address these challenges, we propose AutoMIA, an automated Mirror Illusion Art design pipeline that jointly optimizes shape and color. To stabilize optimization and suppress artifacts, four mechanisms are introduced: (1) projection-alignment component (PAC) selection to reduce surface noise, (2) position-weighted adaptive (PWA) suppression for background noise,  (3) internal voxel preservation (IVP) to prevent internal fractures, and (4) shape-color decoupled (SCD) optimization that balance shape and color optimization. AutoMIA generate diverse smooth Mirror Illusion artworks successfully both in the digital and physical world, with only around 76s design time and 2.6 GB memory on average using a single RTX 3090, advancing inverse graphics and computational design. Our code is available at \url{https://github.com/zxp555/AutoMIA}.
\end{abstract}

\section{Introduction}
\label{sec:intro}

Art and design are important domains for exploring the creativity of AI. In such fields, AI can reduce the amount of complex thinking required by human designers, and even create many unexpected new designs. Among various arts, optical illusion art stands out as both artistically valuable and scientifically insightful, as it exposes the limitations of the human visual system and offers opportunities for deeper understanding of human perception. Existing optical illusion art can be broadly divided into two categories: 2D illusion art and 3D illusion art. Well-known examples include Hybrid Images \cite{oliva2006hybrid} and Visual Anagrams \cite{geng2024visual} in the 2D domain, as well as Shadow Art \cite{mitra2009shadow} and Multi-View Wire Art \cite{hsiao2018multi} in the 3D domain.

In this work, we explore an interesting 3D optical illusion art, termed Mirror Illusion Art. Unlike traditional visual illusions, Mirror Illusion Art leverages the unique properties of reflection: when a 3D object is placed in front of a mirror, its reflection appears as a completely different object, distinct from the original view (see Fig. \ref{fig1}). This phenomenon arises from the deliberate design of both the geometry and texture of the 3D object, thus effectively deceiving human perception. To automate the creation of such illusions, we formulate the following problem: given two 2D images, one representing the appearance in front of the mirror and the other specifying the desired reflection, we develop AI algorithms to generate a 3D illusion object whose real-world reflection achieves the target Mirror Illusion effect.

Sugihara \cite{sugihara2018topology} proposed a topology-based design method for Mirror Illusion Art. However, this approach has two key limitations. First, the design process heavily relies on human intuition and intricate mathematical calculations, making it challenging for beginners. Second, the method focuses solely on shape optimization and does not support the creation of designs involving colored patterns. Another approach is based on Shadow Art \cite{mitra2009shadow}, however, this method also restricts optimization to shape and cannot handle color patterns. Moreover, since Shadow Art primarily focuses on the projected shadow rather than the geometry of the 3D object itself, it often results in 3D objects that are not smooth or complete, which in turn degrades the visual quality and physical realizability of Mirror Illusion Art.

To address these limitations, we propose an automated method named AutoMIA for designing 3D Mirror Illusion Art. Given any pair of user-specified 2D images, our approach optimizes a 3D model to produce the desired mirror illusion art. Furthermore, our method jointly optimizes both 3D shape and color, enabling the creation of smooth and color-rich 3D Mirror Illusions. 
To attain this goal, we found challenges to remove several annoying defects, including ``surface nosie'', ``background noise'', ``internal fracture'' and ``color-shape imbalance''.

To address these challenges, we introduce several targeted solutions. We propose a projection alignment-based component (PAC) selection method to mitigate the ``surface noise''. To suppress ``background noise'', we develop an position-weighted adaptive (PWA) noise suppression method. For resolving ``internal fracture'' inside the 3D object, we design an internal voxel preservation (IVP) mechanism during optimization. Finally, to alleviate the ``color-shape imbalance'', we develop a shape-color decoupled (SCD) optimization method.

Experiments demonstrate that our method can efficiently generate various Mirror Illusion Arts, covering diverse categories such as letters, digits, Chinese characters, cartons, logos, emoji icons, and so on. Moreover, all of our designs are physically realizable, Figure \ref{fig1} showcases real-world artworks produced using our method. On average, our approach requires only around 76 seconds of design time and 2.6 GB of memory on average using a single RTX 3090 GPU, highlighting its efficiency and lightweight advantage. This work shows the synergy of science and art, highlighting how AI can help create new kinds of art and also help us better understand how people see and interpret visual patterns.

\begin{figure*}[htbp]
\centering
\includegraphics[width=1.8\columnwidth]{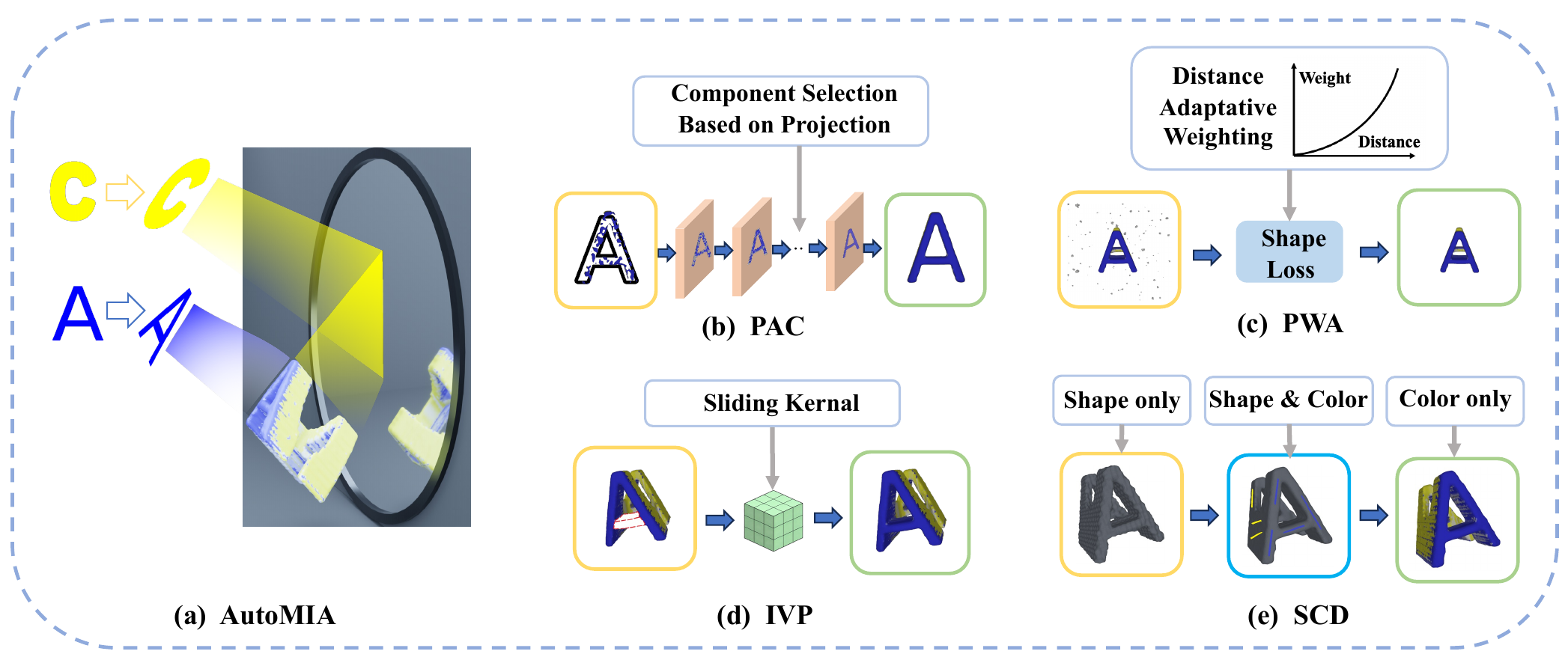} 
\caption{Pipeline of the proposed method. (a) Automatic design of Mirror Illusion Art (AutoMIA). (b) Projection Alignment-Based Component Selection (PAC). (c) Position-Weighted Adaptive Noise Suppression (PWA). (d) Internal Voxel Preservation Mechanism (IVP). (e) Shape-Color Decoupled Optimization (SCD). }
\label{main_process}
\end{figure*} 

\section{Related Work}

\subsection{2D Illusion Arts}

Early computational approaches to 2D illusion art exploit specific properties of human vision. For example, hybrid images \cite{oliva2006hybrid} utilize multi-scale perception by combining low- and high-frequency components from two images, producing a morphing effect with viewing distance. Motion illusions \cite{freeman1991motion} leverage the visual system’s sensitivity to local phase changes, generating continuous motion perception without actual displacement. Camouflage arts \cite{chu2010camouflage,owens2014camouflaging,guo2023ganmouflage,zhang2020deep} hide targets by matching background statistics and adjusting luminance or feature cues, often using learning-based objectives for naturalness and difficulty control. However, these illusion arts are time-consuming to design, and usually require a high level of expertise. 


Recent approaches frame 2D illusion synthesis using generative priors. Diffusion-based methods \cite{geng2024visual,burgert2024diffusion} generate multi-view illusions by jointly denoising across pixel transformations, enabling concealment without explicit perceptual models. Analyses of GAN imagery \cite{hertzmann2020visual} highlight “visual indeterminacy” as a source of illusion-like artifacts. However, these methods are computationally expensive and limited by the inherent randomness of generative models, making it difficult to produce stable illusion patterns.

\subsection{3D Illusion Arts}


Compared to 2D illusion arts, 3D illusion arts involve more complex geometric relationships and must maintain illusion consistency across multiple viewpoints, distances, and lighting conditions, making both design and realization more challenging. Common types include shadow-arts \cite{mitra2009shadow,sadekar2022shadow,gangopadhyay2025hand,min2017soft}, which inverse-design 3D sculptures to match target shadows from specific light directions; wire-arts \cite{hsiao2018multi,qu2024wired}, which construct connected 3D wire networks reproducing line drawings from multiple views; and warp-arts \cite{feng2024illusion3d,chang2025lookingglass}, which create multi-view or anamorphic illusions by learning viewpoint-dependent warps to reveal hidden patterns.

Mirror Illusion Art is an interesting 3D illusion art that optimizes a special 3D object capable of presenting completely different patterns in front of and inside a mirror. Sugihara \cite{sugihara2018topology} proposed a topology-based design method for Mirror Shadow Art. However, this approach heavily relies on human intuition and math, and it only supports shape optimization. Although some shadow art methods \cite{mitra2009shadow,sadekar2022shadow} can be modified to achieve a certain level of effect, these methods also restrict optimization to shape and often result in 3D objects that are not smooth or complete.

\section{Method}

\subsection{Problem Formulation and Overview}

Our goal is to automatically reconstruct a 3D object $V$ that exhibits a ``mirror illusion'' effect, supervised by two given colored images, $A$ and $B$. Specifically, we require the 3D object’s front view ($C_\mathrm{direct}$) in front of a mirror appears as $A$, while its mirrored view ($C_\mathrm{mirror}$) resembles $B$. We denote the 3D renderer as $\mathrm{R}$, and the viewing angles in front of and within the mirror as $\theta_\mathrm{direct}$ and $\theta_\mathrm{mirror}$, respectively. The similarity function $\Phi$ measures the resemblance between the rendered view and the target image, incorporating both shape and color similarity. The optimization objective is thus given by:

\begin{equation}\label{eq:goal}
    \begin{gathered}
    \min_V\!\bigl(\Phi(C_{\mathrm{direct}},A)+\Phi(C_{\mathrm{mirror}},B)\bigr)=\\
    \min_V\!\bigl(\Phi(\mathrm{R}(V,\theta_{\mathrm{direct}}),A)+\Phi(\mathrm{R}(V,\theta_{\mathrm{mirror}}),B)\bigr).
    \end{gathered}
\end{equation}

The function $\Phi$ consists of the shape loss function $L_{\text{shape}}$ and the color loss function $L_{\text{color}}$. $L_{\text{shape}}$ measures the Binary Cross Entropy (BCE) loss between the target pattern mask and the projected mask, while $L_{\text{color}}$ measures the $L_1$ loss between the target pattern’s color values and the projected pattern’s color values. Specifically, suppose there are $N$ pixels on the current projection plane $F$. For the $i$-th pixel ($i \in [1, N]$), let $m_i^t$ and $c_i^t$ denote the mask value and color value of the target pattern, respectively, and $m_i^r$ and $c_i^r$ denote the corresponding mask value and color value of the rendered projection. Then

\begin{equation} \label{eq:l_shape}
    L_\text{shape}=-\frac{1}{N}\sum_{i=1}^{N}(m_{i}^{t}\cdot log(m_{i}^{r})+(1-m_{i}^{t})\cdot\log{(1-m_{i}^{r})}),
\end{equation}
and
\begin{equation} \label{color_loss}
    L_\text{color}=-\frac{1}{N}\sum_{i=1}^N|c_i^t-c_i^r|.
\end{equation}

To facilitate the optimization in 3D space, we build a 3D voxel model of the object $V$, represented as
\begin{equation}
    V=\{(x_i,\rho_i,c_i)\},\quad i=1,2,\ldots,L,
\end{equation}
where $L$ denotes the total number of voxels in the 3D object $V$, $x_i$ is the 3D spatial coordinate of each voxel, $\rho_i \in [0, 1]$ indicates the density (opacity) of each voxel, and $c_i$ represents the color of each voxel. We jointly optimize $x_i$, $\rho_i$, and $c_i$ to achieve our goal (Equation \ref{eq:goal}).


However, during optimization, we identified four main challenges: ``surface noise'', ``background noise'', ``internal fracture'', and ``color-shape imbalance''. We address these with four corresponding methods: Projection-Alignment Component (PAC) selection (Section \ref{sec:PAC}), Position-Weighted Adaptive (PWA) suppression (Section \ref{sec:PWA}), Internal Voxel Preservation (IVP) (Section \ref{sec:IVP}), and Shape-Color Decoupling (SCD) optimization (Section \ref{sec: SCD}).


After that, we convert the optimized 3D voxel model into a smoother 3D mesh model and fabricate it using 3D printing (detailed in \textit{Supplementary Material (SM)}). Figure \ref{fig1} presents the overall pipeline of our method.

\begin{figure}[tbp]
\centering
\includegraphics[width=0.8\columnwidth]{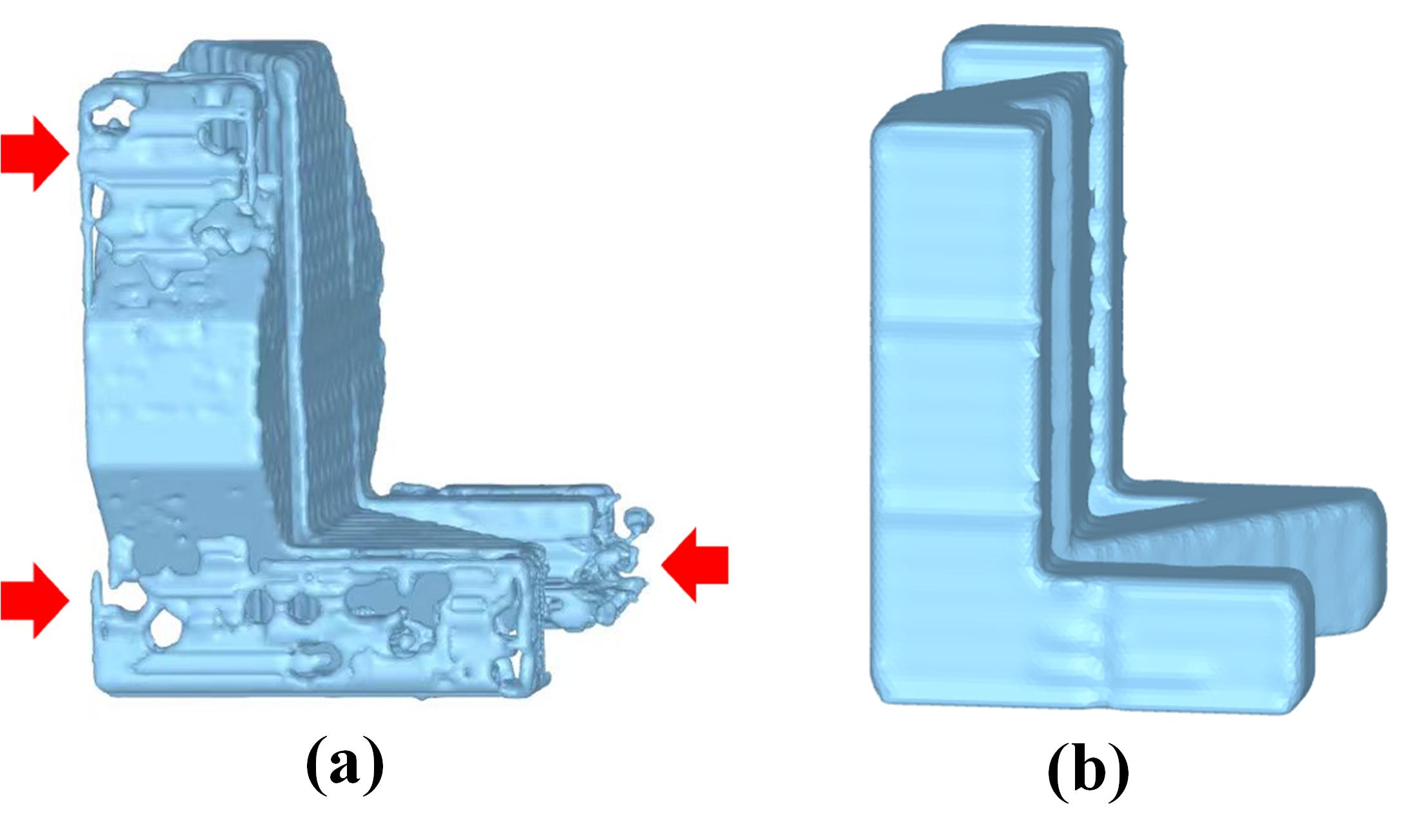} 
\caption{ The ``surface noise'' defect (a)  and its mitigation after applying our PAC method (b). }
\label{surface noise}
\end{figure} 

\subsection{Projection Alignment-Based Component Selection} \label{sec:PAC}

Since the 3D object is constrained by two different view patterns during the reconstruction process, the supervision signal from one angle may introduce noise into the reconstruction from the other angle. This often results in surface voxel noise on the 3D object, which in turn affects the quality of the projected views. We refer to this as the ``surface noise''. Figure \ref{surface noise}(a) shows an example.

To address this challenge, we propose a projection alignment-based component (PAC) selection method. A component is defined as an independent subset of spatially connected voxels. Two voxels are considered connected if they share a common face. We use a depth-first traversal algorithm to identify all $K$ components in the current 3D space, denoted as $S_k$, $k=1,2,\ldots,K$. Next, for any given component $S_k$, we render its projections under $\theta_\mathrm{direct}$ and $\theta_\mathrm{mirror}$ to obtain the projection masks $M_k^\mathrm{direct}$ and $M_k^\mathrm{mirror}$, respectively. We then compute their alignment scores with the supervision pattern masks $M_A$ and $M_B$ using the following formula:
\begin{equation}
    \begin{aligned}
l_k &= \alpha\operatorname{IoU}(M_k^\mathrm{direct},M_A)+\beta\operatorname{IoU}(M_k^\mathrm{mirror},M_B) \\
    &\quad -\gamma\bigl(O(M_k^\mathrm{direct},M_A)+O(M_k^\mathrm{mirror},M_B)\bigr).
\end{aligned}
\end{equation}

Here, $\mathrm{IoU}(X, Y) = \frac{|X \cap Y|}{|X \cup Y|}$ measures the overlap between each component’s projection mask and the corresponding supervision mask, while $O(X, Y) = \frac{|X \cap (1 - Y)|}{|X|}$ quantifies the proportion of pixels in $X$ that fall outside the supervision mask $Y$. $\alpha$, $\beta$, and $\gamma$ are weighting coefficients. We set a threshold $\tau$ for the alignment score $l_k$, and retain only those components with $l_k \ge \tau$, removing all others.

Through this approach, we iteratively remove surface noise from the 3D object $V$ while preserving its effective main structure. This results in a smoother object surface. Figure \ref{surface noise} (b) presents an example after denoising.

\subsection{Position-Weighted Adaptive Noise Suppression} \label{sec:PWA} 

\begin{figure}[tbp]
\centering
\includegraphics[width=0.8\columnwidth]{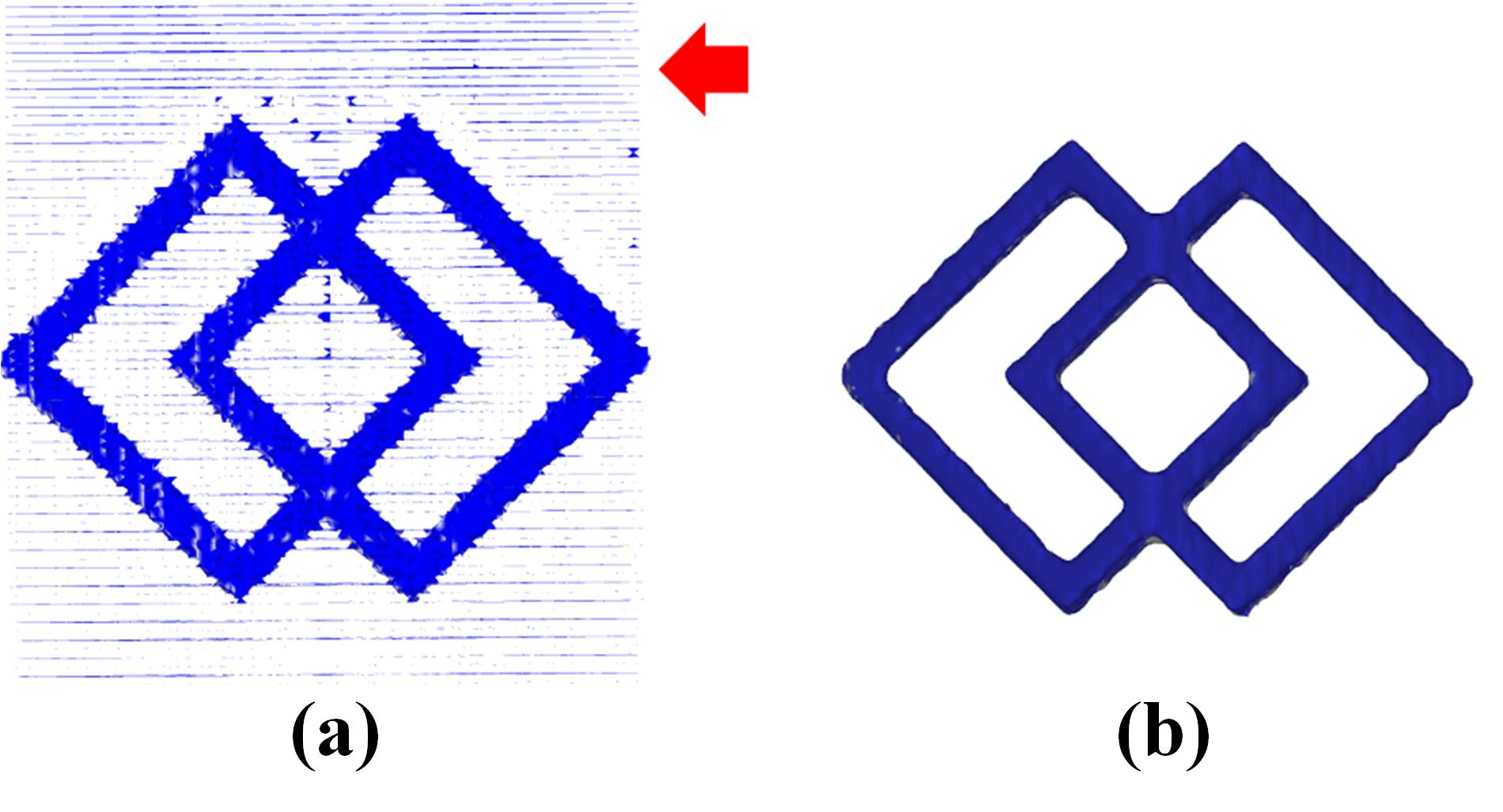} 
\caption{ The ``background noise'' defect (a)  and its mitigation after applying our PWA method (b). }
\label{background noise}
\end{figure} 

Since the supervision images $A$ and $B$ mainly constrain the surface shape and color of the 3D object $V$, their ability to suppress external noise voxels far from the object's surface is limited. As a result, noisy voxels may appear at background locations away from the surface of $V$, as illustrated in Figure \ref{background noise}(a). We refer to this defect as the ``background noise.''

To address this issue, we propose a position-weighted adaptive (PWA) noise suppression method. The core idea is to assign a distance-adaptive weight to each pixel on the projection plane, thereby reducing the loss contribution from pixels that are far from the target mask. In other words, projections that are farther from the target mask are penalized more heavily during optimization. 

Specifically, let $F$ denote the projection plane in the viewing direction. Both the projection mask $C$ of the 3D object $V$ and the shape mask $M$ of the target image are defined on this plane. For any pixel $u$ on $F$, we let $d_{\text{max}}(u)$ denote the maximum Euclidean distance from $u$ to the farthest pixel in $M$. Similarly, $d_{\text{min}}(u)$ denotes the minimum Euclidean distance from $u$ to the nearest pixel in $M$. We then define the distance-adaptive weight $w(u)$ as follows:
\begin{equation}
    w(u)=1+(w_{max}-1)\cdot\left(\frac{d_{min}(u)}{d_{max}(u)}\right)^q
\end{equation}
Here, $q$ is a hyperparameter that controls the distance gain, and the maximum value of $w(u)$ is set to $w_{max}$. Based on Equation \ref{eq:l_shape}, the shape optimization loss $L_{\text{shape}}$ after incorporating the distance-adaptive weighting is given by:
\begin{equation}
\begin{aligned}
L_\text{shape} &= -\frac{1}{N}\sum_{i=1}^N w(u_i)\cdot\bigl(m_i^t\cdot\log(m_i^r)\\
          &\quad +(1-m_i^t)\cdot\log\bigl(1-m_i^r\bigr)\bigr).
\end{aligned}
\end{equation}

In this way, we can effectively penalize projection pixels that are in the background areas, with the penalty increasing as the distance grows. This approach helps eliminate distant external noise. Figure \ref{background noise}(b) presents an example after applying this method.

\begin{figure}[tbp]
\centering
\includegraphics[width=0.6\columnwidth]{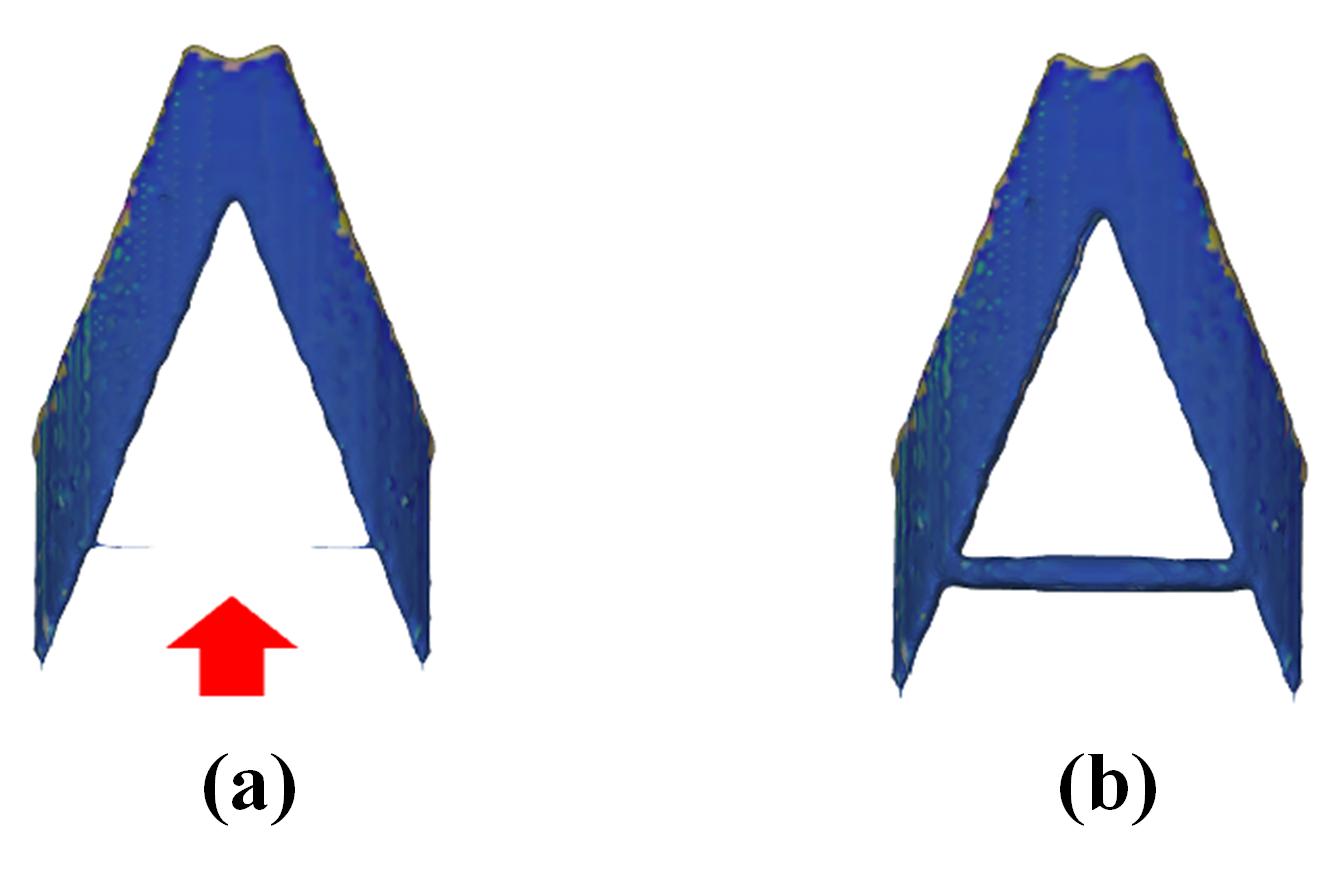} 
\caption{ The ``internal fracture'' defect (a)  and its mitigation after applying our IVP method (b). }
\label{internal fracture}
\end{figure} 

\subsection{Internal Voxel Preservation Mechanism} \label{sec:IVP}

Since the two supervision images A and B mainly constrain the optimization of surface voxels of the 3D object $V$, the voxels located inside $V$ lack direct supervision signals. We observe that some internal voxels may have their densities reduced to zero (i.e., they disappear) during optimization and are difficult to recover. This leads to internal fractures within the 3D object $V$, preventing it from forming a complete and connected structure—a defect we refer to as ``internal fracture'', as illustrated in Figure \ref{internal fracture}(a).

To address this challenge, we propose an internal voxel preservation (IVP) mechanism. Specifically, we first define the ``solid voxel'' as voxels whose density $\rho(x)$ exceeds a threshold $\gamma$. The identification rule is expressed as $o(x) = \mathbf{1}[\rho(x) > \gamma]$, where $o(x) = 1$ indicates that $x$ is a ``solid voxel''. 

After that, we design a kernel $\Omega$ of size $k \times k \times k$ to determine whether a voxel belongs to the internal region. The core idea is to slide the kernel $\Omega$ over all voxels in the 3D object. For each voxel at the center of $\Omega$, it is classified as an “internal voxel” only if all other voxels within the kernel are ``solid voxel''. This rule can be formulated as follows:

\begin{equation}
    m(x)=\mathbf{1}\!\left[\frac{1}{|\Omega|}\sum_{y\in \Omega(x)} o(y)=1\right].
\end{equation}

When $m(x) = 1$, $x$ is identified as an ``internal voxel''. Based on this mechanism, we impose a lower bound constraint on the density of all ``internal voxel'', requiring their density to be no less than $\rho_{\min}$. This approach ensures that the density of ``internal voxels'' does not decrease to zero, thereby preventing the ``internal fracture'' defect and maintaining the connectivity and integrity of the 3D object $V$. This is also beneficial for subsequent 3D printing. Figure \ref{internal fracture}(b) presents an example after applying this method.

\subsection{Shape-Color Decoupled Optimization} \label{sec: SCD}

\begin{figure}[htbp]
\centering
\includegraphics[width=0.7\columnwidth]{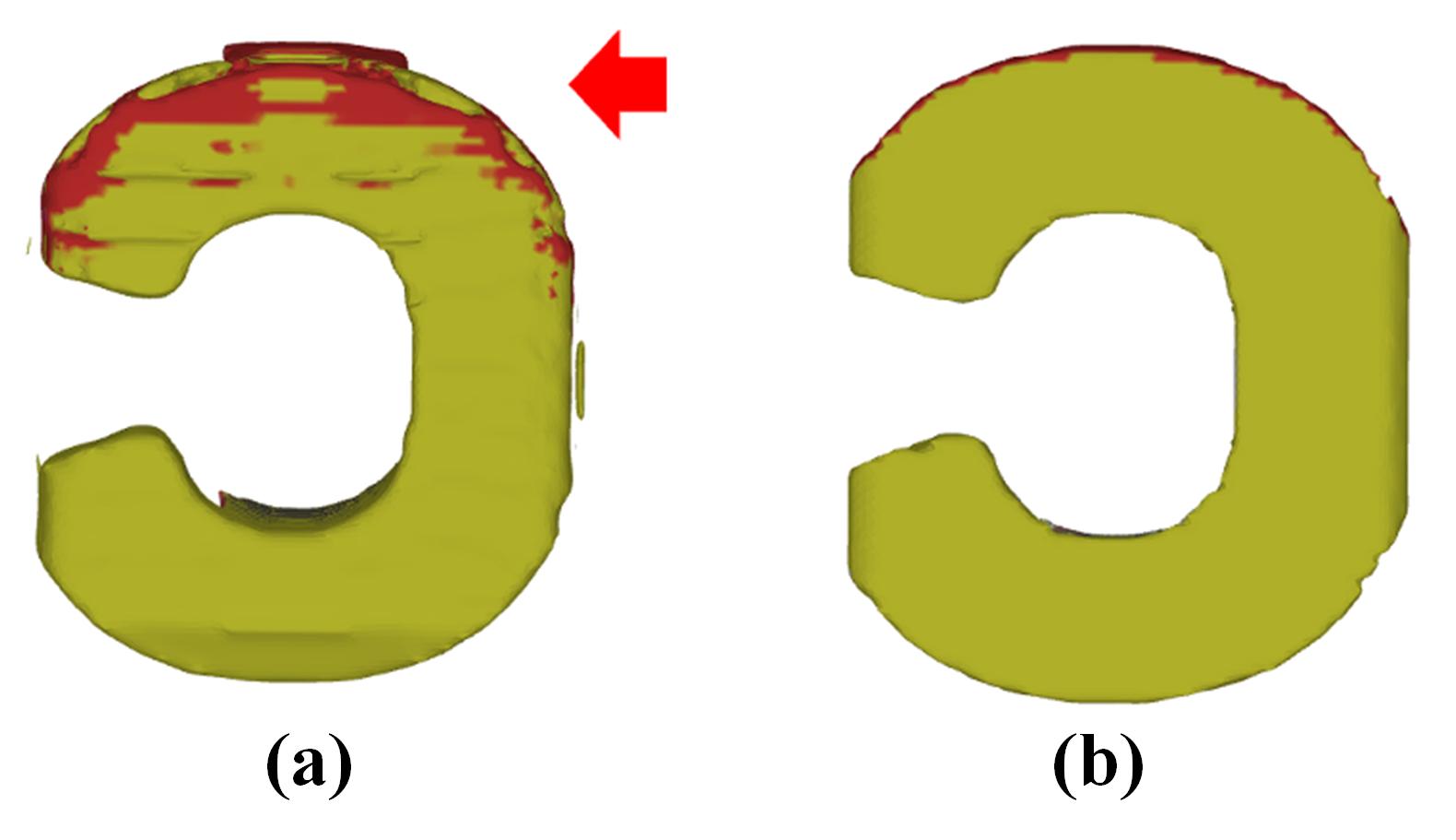} 
\caption{ The ``color-shape imbalance'' defect (a)  and its mitigation after applying our SCD method (b). The target view is a yellow letter C. In sub-figure (a), although the shape meets the requirement, red color from the other view ``leaks'' into the current view, resulting in a color inconsistency. }
\label{color-shape imbalance}
\end{figure}

Since our approach jointly optimizes both the surface shape and color of the 3D object, balancing these two objectives is a critical challenge. We observe that the optimization process often exhibits a ``color-shape imbalance'' defect, where the optimized 3D object $V$ may satisfy only the shape requirement or only the color requirement, but not both, as illustrated in Figure \ref{color-shape imbalance}(a).

To address this issue, we propose a shape-color decoupled (SCD) optimization method. Specifically, we divide the optimization timeline $[0,T]$ into three stages. In the first stage ($0 \leq t < t_1$), SCD optimizes only the shape of the 3D object, ensuring a stable initial structure. In the second stage ($t_1 \leq t < t_2$), SCD jointly optimizes both the shape and color for refinement. In the third stage ($t_2 \leq t \leq T$), SCD optimizes only the color to fine-tune the appearance. The overall loss can thus be expressed as
\begin{equation}
    L_{\mathrm{total}}=w_\mathrm{s}(t)\cdot L_{\mathrm{shape}}+w_\mathrm{c}(t)\cdot L_{\mathrm{color}},
\end{equation}
where $w_\mathrm{s}(t)$ and $w_\mathrm{c}(t)$ are the weights that balance shape and color optimization, respectively, and are defined as follows:
\begin{equation}
w_{\mathrm{s}}(t)=
\begin{cases}
1, & 0 \le t < t_2,\\
0, & t_2 \le t \le T.
\end{cases}
\end{equation}
and
\begin{equation}
w_{\mathrm{c}}(t)=
\begin{cases}
0,      & 0 \le t < t_1,\\
\lambda, & t_1 \le t < t_2,\\
1,      & t_2 \le t \le T.
\end{cases}
\end{equation}
where $\lambda \in (0, 1)$ is the weighting factor that balances $L_{\text{shape}}$ and $L_{\text{color}}$. Through this staged optimization approach, we are able to effectively balance the shape and color optimization of the 3D object, resulting in mirror illusion art that meets both shape and color requirements. Figure \ref{color-shape imbalance}(b) presents an example after applying this method.

\section{Experiments}

\subsection{Datasets}

To show the generality of our method, we collected a wide variety of 2D images as input for our models. These images include English letters, Arabic numerals, Chinese characters, geometric patterns, emoji icons, cartoon designs, and commercial logos, etc.
For each category, we randomly sample 200 images, resulting in a total of 1200 images, similar to the settings used in previous works \cite{mitra2009shadow,sadekar2022shadow,geng2024visual,chang2025lookingglass,feng2024illusion3d}.
The images were either generated using Python scripts or collected from the internet, and all internet-sourced images comply with their respective copyright agreements and are used solely for academic research. We name this dataset \textit{Mirror-2D} and will release it soon.

\subsection{Evaluation Metrics}
We employ the following evaluation metrics to systematically assess the 3D generation quality of different methods. Shape Score (SS) evaluates the shape consistency between the 3D object's projections and the supervision images, ranging from $0$ to $1$, with higher values indicating greater shape similarity. Color Score (CS) assesses color consistency, ranging from $0$ to $1$, with lower values indicating greater color similarity. Noise Level (NL) measures the intensity of surface noise in the 3D voxel model, ranging from $0$ to $1$, with lower values indicating less noise. Smooth Level (SL) evaluates the surface smoothness of the 3D mesh model, ranging from $0$ to $1$, with higher values indicating a smoother surface. The formula definitions and computation details are provided in the \textit{SM}.

\subsection{Experimental Settings} \label{sec:setting}

To ensure fair comparison, we adopt the same experimental settings for all experiments. The resolution of the 3D voxel model is set to $128 \times 128 \times 128$. Rendering is performed using PyTorch3D \cite{ravi2020accelerating}.
See \textit{SM} for more details.

\subsection{Baselines}

We select the methods of Shadow Art (SA) \cite{mitra2009shadow} and Shadow Art Revisited (SAR) \cite{sadekar2022shadow}  as baselines.
Since both SA and SAR support only shape optimization and do not allow for color optimization, we convert the input 2D color images to 2D masks during optimization, with other settings following Section \ref{sec:setting}. We use only SS, NL, and SL as evaluation metrics when comparing our method with these baselines for the same reason. Since we primarily focus on automated design methods and the results of manual design are highly influenced by individual designer differences, which makes quantitative evaluation challenging, therefore, we do not include the manual design approach \cite{sugihara2018topology} as baselines in this work.


\begin{table}[htbp]
\centering
\small
\caption{Comparision with baseline methods}\label{compare}
\begin{tabular}{c|ccccc}
\toprule[1.1pt]  
Methods & SL~$\uparrow$ & NL~$\downarrow$ & SS~$\uparrow$ & Time~$\downarrow$ & Memory~$\downarrow$ \\
\hline
SA \cite{mitra2009shadow} & 0.827 & 0.507 & 0.499  & 50s & 2.5 GB \\
SAR \cite{sadekar2022shadow} & 0.834 & 0.120 & 0.668 & 140s & 3.3 GB \\
Ours & \textbf{0.989} & \textbf{0.049} & \textbf{0.931} & 76s & 2.6 GB \\
\bottomrule[1.1pt] 
\end{tabular}
\end{table}



\begin{figure}[htbp]
\centering
\includegraphics[width=1\columnwidth]{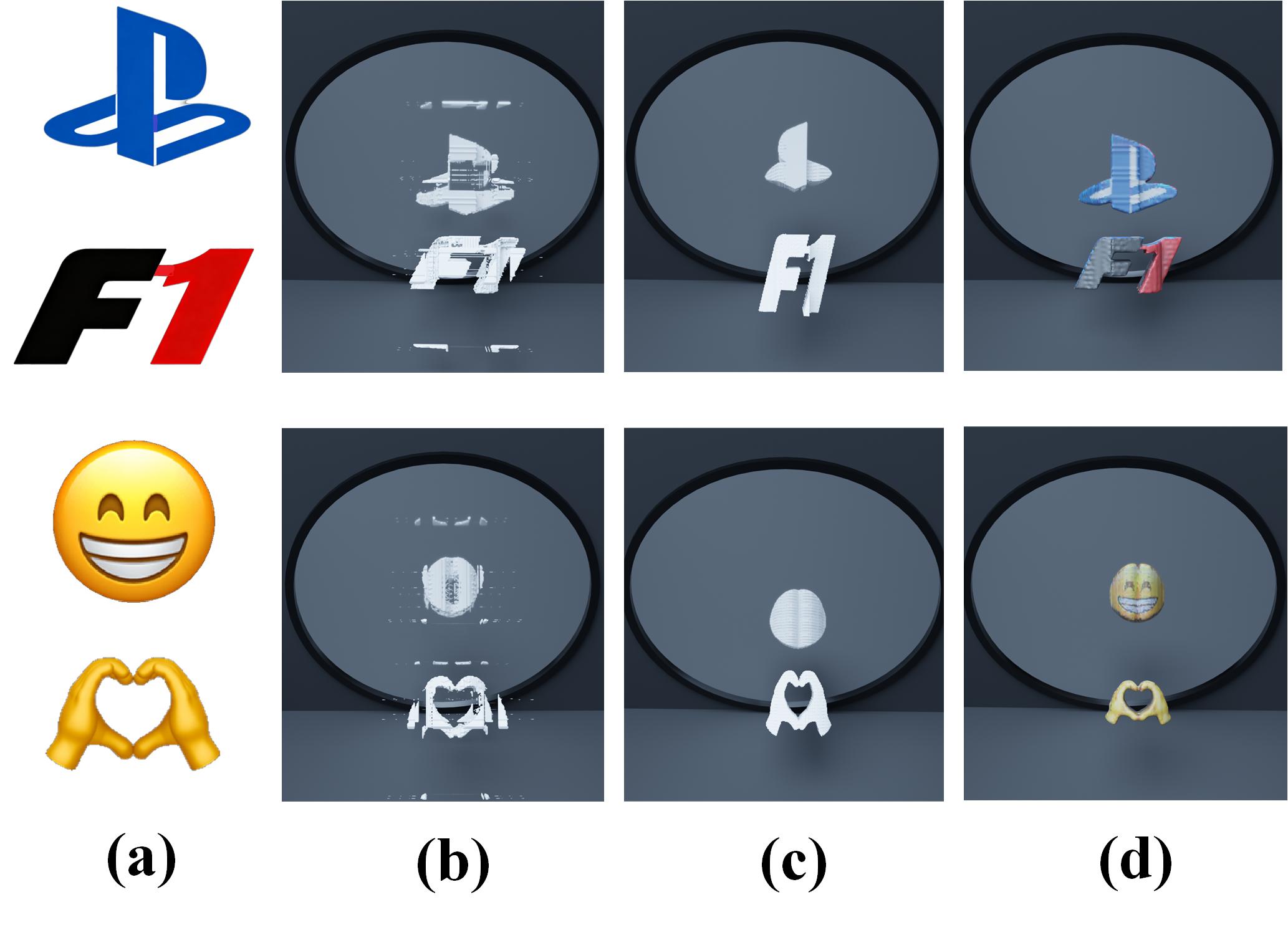} 
\caption{ Examples of 3D objects generated by different methods. (a) Input 2D images. (b) Shadow Art (SA). (c) Shadow Art Revisited (SAR). (d) AutoMIA (Ours).}
\label{visual_compare}
\end{figure}

\begin{figure*}[htbp]
\centering
\includegraphics[width=2\columnwidth]{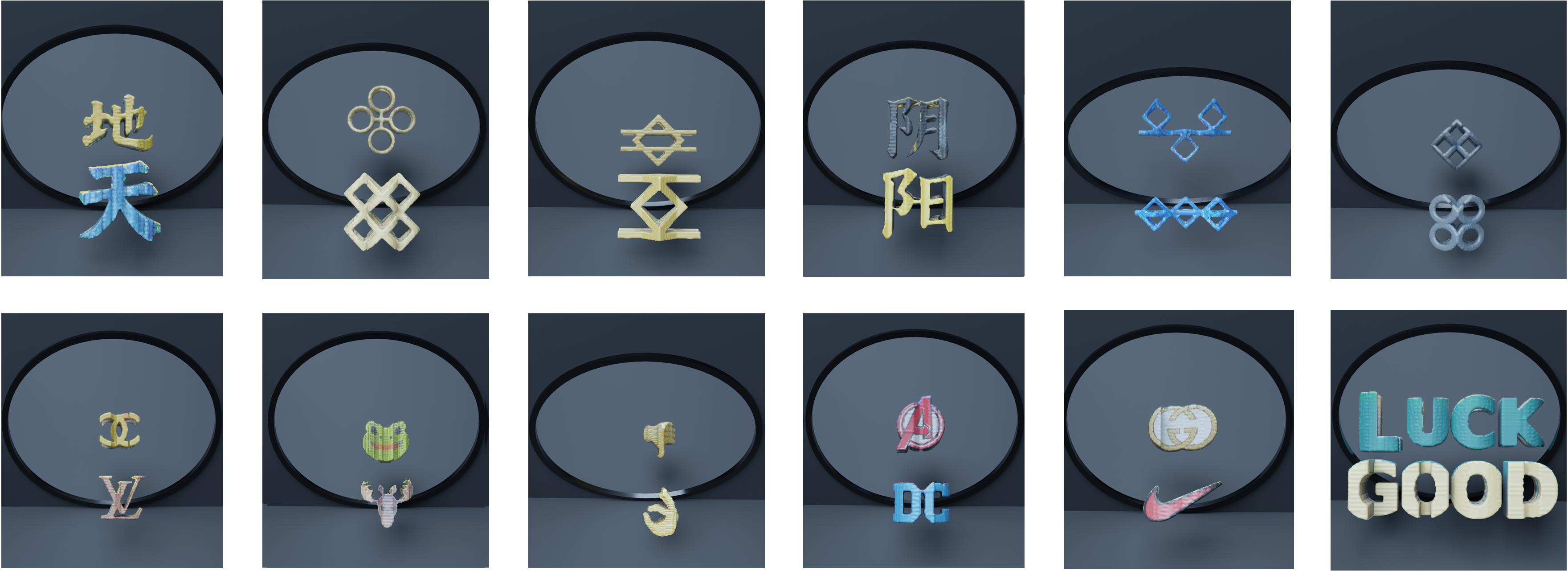} 
\caption{ Digital visualization of Mirror Illusion Arts designed by our AutoMIA method. See \textit{SM} for more examples.}
\label{digital_visual}
\end{figure*}

\subsection{Comparison with Baseline Methods}


We evaluated the performance of our method and the baseline methods (SA \cite{mitra2009shadow} and SAR \cite{sadekar2022shadow}) on the Mirror-2D dataset. The reconstruction quality was assessed using SS, NL, and SL. The results are shown in Table \ref{compare}. The results indicate that our method achieved the best performance across all three metrics. A set of visualization examples is presented in Figure \ref{visual_compare}. As can be seen, the SA and SAR methods not only achieved lower consistency with the target patterns, but also exhibited defects such as surface noise, background noise, and internal fractures. In contrast, our approach produced 3D objects that are closer to the target patterns, with smoother surfaces and less noise, thus achieving the desired Mirror Illusion Art effect.
In addition to shape reconstruction, our method can also accurately recover the color patterns of the 2D images, whereas the two baseline methods are limited to shape reconstruction only. This highlights a unique advantage of our approach.

The main reasons are as follows. First, SA and SAR are mainly designed for orthogonal settings (e.g. reconstructing 3D objects from 3 orthogonal images) and struggle with non-orthogonal scenarios like Mirror Illusion Art (reconstructing 3D objects from 2 non-orthogonal images), where limited supervision leads to more surface noise and unwanted structures. In contrast, our proposed PAC and PWA effectively suppress such noise under non-orthogonal settings, demonstrating their unique advantages. Second, both SA and SAR do not account for internal connectivity, often resulting in internal fractures. In contrast, our IVP method effectively prevents the occurrence of ``internal fractures'', ensuring the integrity of the optimized 3D patterns and making them more suitable for physical fabrication. Third, compared to SA and SAR, we introduce color constraints and propose the SCD method to balance shape and color optimization, enabling simultaneous reconstruction of both shape and color.

We further evaluated the average computational time and memory consumption on the same RTX 3090 GPU. The results (Table \ref{compare}) show that, despite incorporating 4 core modules to enforce physical constraints, AutoMIA achieves similar computational speed and resource usage to SA and is significantly more efficient than SAR, while delivering substantially better reconstruction quality than both baselines. This demonstrates the combined advantages of our method in both efficiency and reconstruction quality.


\begin{table}[bp]
\centering
\small
\caption{Ablation Study}\label{ablation}
\begin{tabular}{c|cccc}
\toprule[1.1pt]  
Methods & SL~$\uparrow$ & NL~$\downarrow$ & SS~$\uparrow$ & CS~$\downarrow$ \\
\hline
Ours & \textbf{0.989} & \textbf{0.049} & \textbf{0.931} &  \textbf{0.018} \\
$-$ PAC & 0.910 & 0.248 & 0.629 & 0.323 \\
$-$ PWA & 0.822 & 0.373 & 0.790 & 0.034 \\
$-$ IVP & 0.979 & 0.050 & 0.748 & 0.021 \\
$-$ SCD & 0.755 & 0.101 & 0.548 & 0.050 \\
\bottomrule[1.1pt] 
\end{tabular}
\end{table} 

\subsection{Digital Visualization} \label{sec:digital visual}

We constructed a virtual environment for digital visualization of Mirror Illusion Art, based on Blender 4.5. See \textit{SM} for the details. 
Figure \ref{digital_visual} presents several examples of digital Mirror Illusion Art visualizations, with additional examples provided in the \textit{SM}. Interestingly, when viewing these designs for the first time, the mirrored pattern often appears quite different from the pattern seen in front of the mirror, making it seem as if they are two distinct objects rather than a single object and its mirror view. We provide an in-depth discussion of the underlying neuroscience and psychological mechanisms behind this illusion phenomenon in the \textit{SM}.


\subsection{Ablation Study}

PAC, PWA, IVP, and SCD are the four core modules of our method. We conducted an ablation study to demonstrate their individual contributions, with the results summarized in Table \ref{ablation}. The findings indicate that removing any one of these modules leads to a noticeable decline in 3D reconstruction quality. Specifically, PAC has a significant impact on NL, SS, and CS; PWA mainly affects SL, NL, and SS; IVP primarily influences SS; and SCD shows a clear effect on SL, SS, and CS. Interestingly, adding any module improves all four metrics, even if a module is mainly designed for specific aspects. For example, while PWA primarily targets SL, NL, and SS, it also brings positive effects for CS. This may be because our method effectively balances shape and color optimization, and smoother surfaces with less noise are beneficial for further color refinement. Figures \ref{surface noise}-\ref{color-shape imbalance} provide visual examples to intuitively illustrate the effects of each module.


\subsection{Volume Size and Ray Sampling Density}

During optimization, the volume size and ray sampling density are two important hyperparameters that directly affect both the quality of the rendered 3D object and the computational cost. We provide a detailed quantitative analysis of these parameters in \textit{SM}.

\subsection{Physical Visualization}

We selected six representative 3D objects from different categories and fabricated them using 3D printing technology (See \textit{SM} for details.)
The printed objects were placed in front of a circular mirror with a diameter of 16 cm, and the physical demonstration of Mirror Illusion Art follows similar settings of the simulation environment (Section \ref{sec:digital visual}). The light source was standard indoor lighting, and all photos were taken using an iPhone 13.

Figure \ref{fig1} shows several physical-world visualization examples. Because our optimization process accounts for various physical constraints (removing surface noise, external noise, and internal fractures), our designs can transition smoothly from digital models to physical realizations. In contrast, many 3D patterns generated by SA and SAR cannot be physically fabricated, as these methods do not consider such physical constraints. Especially for 3D objects with intricate line patterns and hollow interiors, SA and SAR tend to produce internal fractures and noise when handling such complex structures, making 3D printing infeasible. In contrast, our method can successfully handle these cases.

\subsection{Discussion}

\begin{figure}[tbp]
\centering
\includegraphics[width=1\columnwidth]{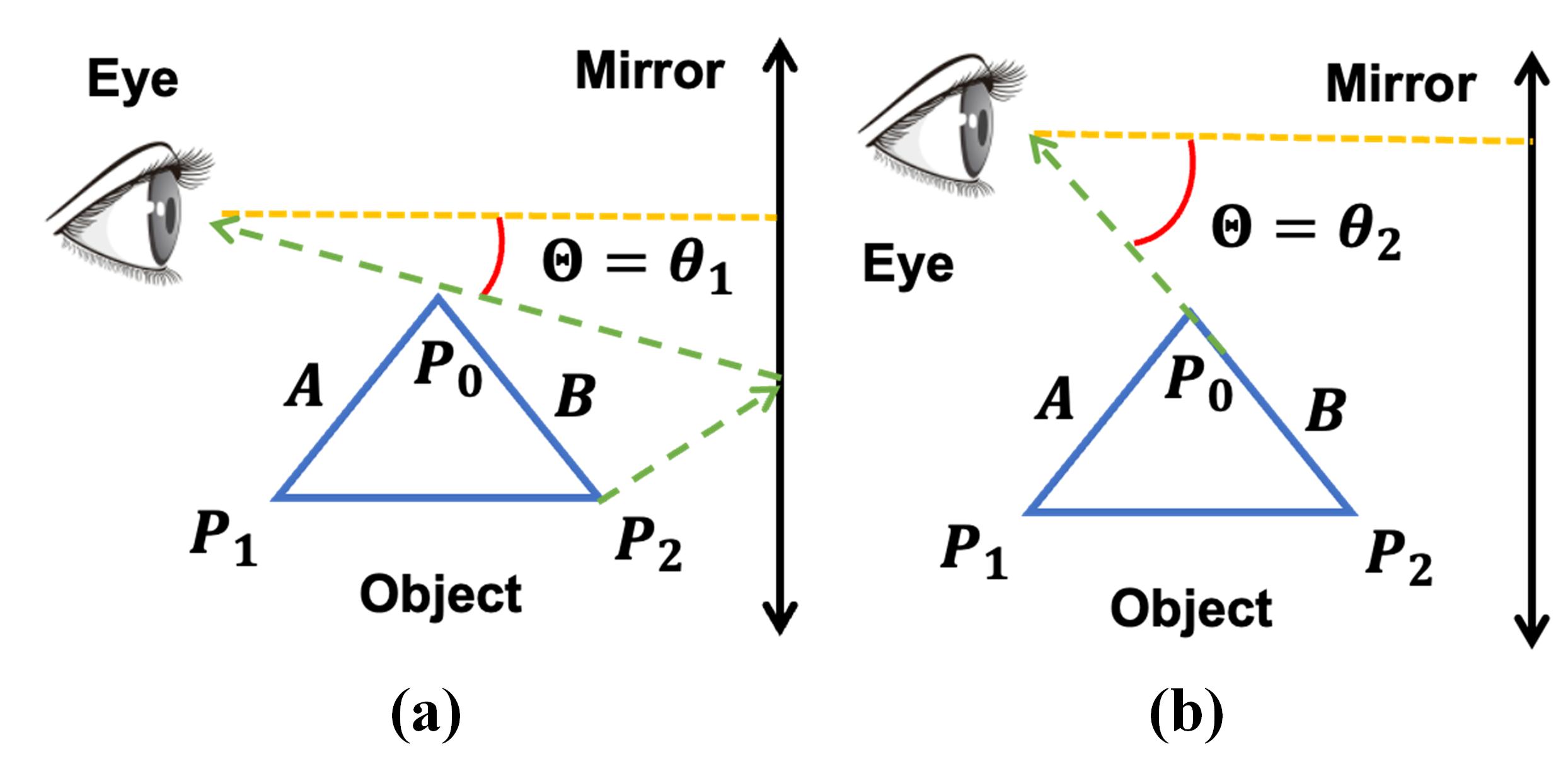} 
\caption{ Mathematical model for Mirror Illusion Art. (a) Boundary case 1. (b) Boundary case 2. }
\label{math}
\end{figure}

Mirror Illusion Art requires specific viewing angles for the illusion to be perceived. An interesting question is: under what conditions can we observe the Mirror Illusion Art effect? To explore this, we constructed a simple mathematical model, as illustrated in Figure \ref{math}. In the figure, A and B represent the front and back of the object, while $P_0$, $P_1$, and $P_2$ denote three vertices of the object.

Next, we consider two boundary cases. In the first case, as shown in Figure \ref{math}(a), a light ray emitted from $P_2$ is reflected by the mirror, passes through $P_0$, and reaches the observer’s eye. In this scenario, the observer can exactly see the entire back pattern B via the mirror. Let $\Theta$ denote the observer’s viewing angle and $\theta_1$ represent the viewing angle for this first case. If $\Theta$ is less than $\theta_1$, the observer will not be able to see the complete back pattern B via mirror reflection. In the second case, as illustrated in Figure \ref{math}(b), a light ray from $P_2$ travels along the surface of the back pattern B, passes through $P_0$, and reaches the eye directly without reflection. In this situation, the observer can directly see the back pattern B, causing the illusion effect to disappear. Let $\theta_2$ denote the viewing angle for this second case. To observe the ideal mirror illusion art, the observer’s viewing angle $\Theta$ should be less than $\theta_2$. In summary, the observer’s viewing angle should satisfy:
\begin{equation}
    \theta_1<\Theta<\theta_2
\end{equation}
where $\theta_1$ and $\theta_2$ are determined by the shape of the object as well as the relative distances between the observer, the object, and the mirror. When this condition is met, the ideal mirror illusion art can be perceived.

\subsection{Limitations}

\begin{figure}[tbp]
\centering
\includegraphics[width=1\columnwidth]{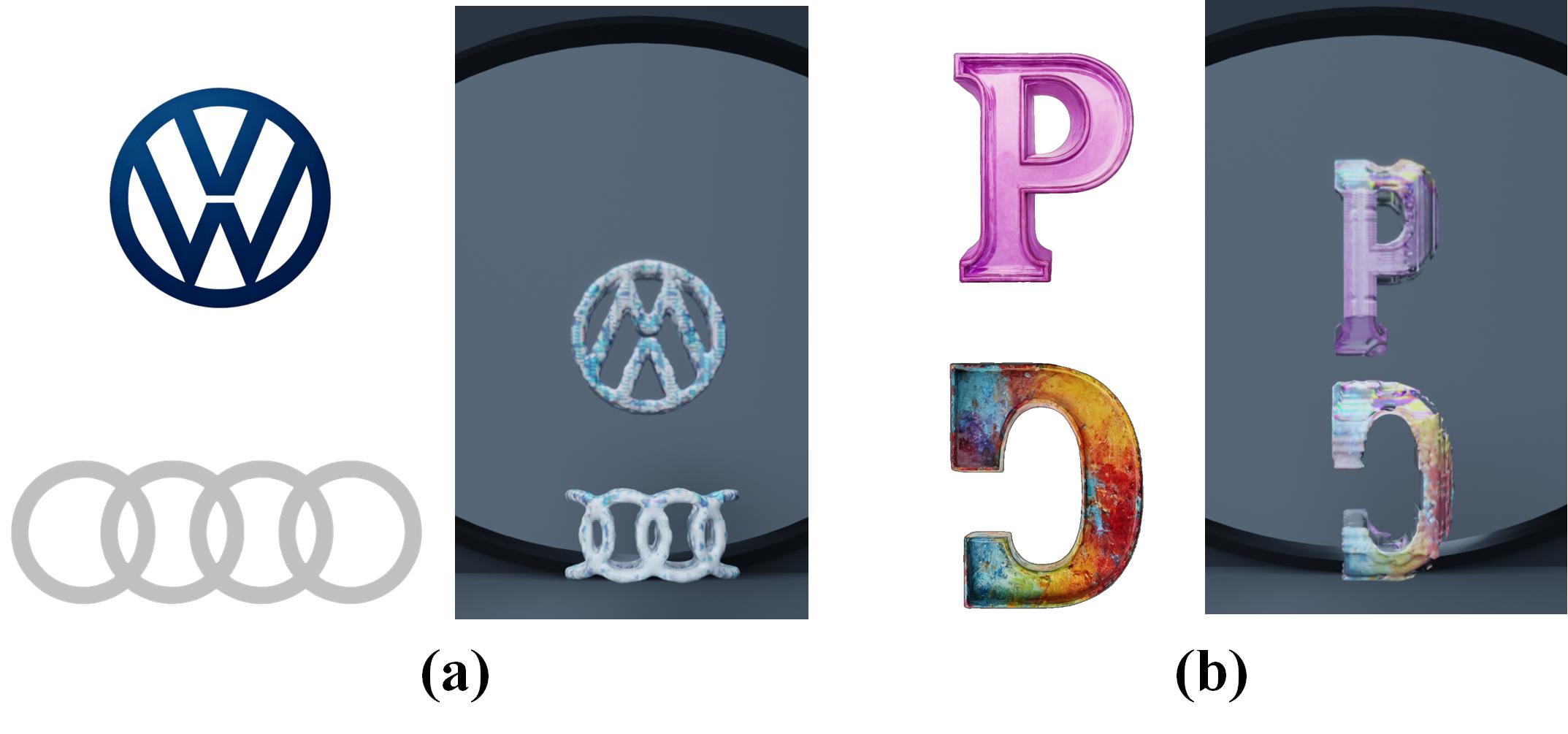} 
\caption{ Typical failure cases. }
\label{failure case}
\end{figure} 




We present AutoMIA for generating diverse and compelling mirror illusion art with low time and computational cost. However, there are two limitations. First, supervision images A and B should have similar widths to avoid conflicting signals and incomplete patterns (Fig. \ref{failure case}(a)), which can be easily resolved by scaling the images to the same width. Second, the image resolution should not exceed the voxel model's resolution, otherwise, high-frequency noise and visual artifacts may occur (Fig. \ref{failure case}(b)). This can be solved by using higher-resolution voxel models (with increased cost) or down-sampling images.

\section{Conclusion}

In this work, we propose AutoMIA, an automated Mirror Illusion Art design pipeline that jointly optimizes both shape and color to generate compelling mirror illusion artworks. To stabilize the optimization and suppress artifacts, we developed the PAC method for reducing surface noise, PWA suppression for background noise, the IVP mechanism to prevent internal fractures, and SCD optimization to balance shape and color optimization. Experimental results demonstrate that AutoMIA successfully produces a wide variety of mirror illusion art in both digital and physical forms, with superior reconstruction quality, fast generation speed, and low computational resource requirements.

\section{Acknowledgments}
This work was supported by the National Natural Science Foundation of China (No. U2341228 and No. 62576187).

{
    \small
    \bibliographystyle{ieeenat_fullname}
    \bibliography{main}

\begin{thebibliography}{25}
\providecommand{\natexlab}[1]{#1}
\providecommand{\url}[1]{\texttt{#1}}
\expandafter\ifx\csname urlstyle\endcsname\relax
  \providecommand{\doi}[1]{doi: #1}\else
  \providecommand{\doi}{doi: \begingroup \urlstyle{rm}\Url}\fi

\bibitem[Beck and Kastner(2009)]{beck2009top}
Diane~M Beck and Sabine Kastner.
\newblock Top-down and bottom-up mechanisms in biasing competition in the human
  brain.
\newblock \emph{Vision research}, 49\penalty0 (10):\penalty0 1154--1165, 2009.

\bibitem[Burgert et~al.(2024)Burgert, Li, Leite, Ranasinghe, and
  Ryoo]{burgert2024diffusion}
Ryan Burgert, Xiang Li, Abe Leite, Kanchana Ranasinghe, and Michael Ryoo.
\newblock Diffusion illusions: Hiding images in plain sight.
\newblock In \emph{ACM SIGGRAPH 2024 Conference Papers}, pages 1--11, 2024.

\bibitem[Chang et~al.(2025)Chang, Sancho, Tang, Gross, and
  Azevedo]{chang2025lookingglass}
Pascal Chang, Sergio Sancho, Jingwei Tang, Markus Gross, and Vinicius Azevedo.
\newblock Lookingglass: Generative anamorphoses via laplacian pyramid warping.
\newblock In \emph{Proceedings of the Computer Vision and Pattern Recognition
  Conference}, pages 24--33, 2025.

\bibitem[Chu et~al.(2010)Chu, Hsu, Mitra, Cohen-Or, Wong, and
  Lee]{chu2010camouflage}
Hung-Kuo Chu, Wei-Hsin Hsu, Niloy~J Mitra, Daniel Cohen-Or, Tien-Tsin Wong, and
  Tong-Yee Lee.
\newblock Camouflage images.
\newblock \emph{ACM Trans. Graph.}, 29\penalty0 (4):\penalty0 51--1, 2010.

\bibitem[Feng et~al.(2024)Feng, Sanjay, Lutz, AlBahar, Ge, and
  Huang]{feng2024illusion3d}
Yue Feng, Vaibhav Sanjay, Spencer Lutz, Badour AlBahar, Songwei Ge, and Jia-Bin
  Huang.
\newblock Illusion3d: 3d multiview illusion with 2d diffusion priors.
\newblock \emph{arXiv preprint arXiv:2412.09625}, 2024.

\bibitem[Freeman et~al.(1991)Freeman, Adelson, and Heeger]{freeman1991motion}
William~T Freeman, Edward~H Adelson, and David~J Heeger.
\newblock Motion without movement.
\newblock \emph{ACM Siggraph Computer Graphics}, 25\penalty0 (4):\penalty0
  27--30, 1991.

\bibitem[Gangopadhyay et~al.(2025)Gangopadhyay, Singh, Tiwari, and
  Raman]{gangopadhyay2025hand}
Aalok Gangopadhyay, Prajwal Singh, Ashish Tiwari, and Shanmuganathan Raman.
\newblock Hand shadow art: A differentiable rendering perspective.
\newblock \emph{arXiv preprint arXiv:2505.21252}, 2025.

\bibitem[Geng et~al.(2024)Geng, Park, and Owens]{geng2024visual}
Daniel Geng, Inbum Park, and Andrew Owens.
\newblock Visual anagrams: Generating multi-view optical illusions with
  diffusion models.
\newblock In \emph{Proceedings of the IEEE/CVF Conference on Computer Vision
  and Pattern Recognition}, pages 24154--24163, 2024.

\bibitem[Guo et~al.(2023)Guo, Collins, de~Lima, and Owens]{guo2023ganmouflage}
Rui Guo, Jasmine Collins, Oscar de Lima, and Andrew Owens.
\newblock Ganmouflage: 3d object nondetection with texture fields.
\newblock In \emph{Proceedings of the IEEE/CVF Conference on Computer Vision
  and Pattern Recognition}, pages 4702--4712, 2023.

\bibitem[Hertzmann(2020)]{hertzmann2020visual}
Aaron Hertzmann.
\newblock Visual indeterminacy in gan art.
\newblock In \emph{ACM SIGGRAPH 2020 Art Gallery}, pages 424--428. 2020.

\bibitem[Hsiao et~al.(2018)Hsiao, Huang, and Chu]{hsiao2018multi}
Kai-Wen Hsiao, Jia-Bin Huang, and Hung-Kuo Chu.
\newblock Multi-view wire art.
\newblock \emph{ACM Trans. Graph.}, 37\penalty0 (6):\penalty0 242, 2018.

\bibitem[Kingma(2014)]{kingma2014adam}
Diederik~P Kingma.
\newblock Adam: A method for stochastic optimization.
\newblock \emph{arXiv preprint arXiv:1412.6980}, 2014.

\bibitem[Knill and Richards(1996)]{knill1996perception}
David~C Knill and Whitman Richards.
\newblock \emph{Perception as Bayesian inference}.
\newblock Cambridge University Press, 1996.

\bibitem[Lorensen and Cline(1998)]{lorensen1998marching}
William~E Lorensen and Harvey~E Cline.
\newblock Marching cubes: A high resolution 3d surface construction algorithm.
\newblock In \emph{Seminal graphics: pioneering efforts that shaped the field},
  pages 347--353. 1998.

\bibitem[Min et~al.(2017)Min, Lee, Won, and Lee]{min2017soft}
Sehee Min, Jaedong Lee, Jungdam Won, and Jehee Lee.
\newblock Soft shadow art.
\newblock In \emph{Proceedings of the symposium on Computational Aesthetics},
  pages 1--9, 2017.

\bibitem[Mitra and Pauly(2009)]{mitra2009shadow}
Niloy~J Mitra and Mark Pauly.
\newblock Shadow art.
\newblock \emph{ACM Trans. Graph.}, 28\penalty0 (5):\penalty0 156, 2009.

\bibitem[Oliva et~al.(2006)Oliva, Torralba, and Schyns]{oliva2006hybrid}
Aude Oliva, Antonio Torralba, and Philippe~G Schyns.
\newblock Hybrid images.
\newblock \emph{ACM Transactions on Graphics (TOG)}, 25\penalty0 (3):\penalty0
  527--532, 2006.

\bibitem[Owens et~al.(2014)Owens, Barnes, Flint, Singh, and
  Freeman]{owens2014camouflaging}
Andrew Owens, Connelly Barnes, Alex Flint, Hanumant Singh, and William Freeman.
\newblock Camouflaging an object from many viewpoints.
\newblock In \emph{Proceedings of the IEEE Conference on Computer Vision and
  Pattern Recognition}, pages 2782--2789, 2014.

\bibitem[Qu et~al.(2024)Qu, Yang, Zhang, Xiang, Pang, and Song]{qu2024wired}
Zhiyu Qu, Lan Yang, Honggang Zhang, Tao Xiang, Kaiyue Pang, and Yi-Zhe Song.
\newblock Wired perspectives: Multi-view wire art embraces generative ai.
\newblock In \emph{Proceedings of the IEEE/CVF Conference on Computer Vision
  and Pattern Recognition}, pages 6149--6158, 2024.

\bibitem[Ravi et~al.(2020)Ravi, Reizenstein, Novotny, Gordon, Lo, Johnson, and
  Gkioxari]{ravi2020accelerating}
Nikhila Ravi, Jeremy Reizenstein, David Novotny, Taylor Gordon, Wan-Yen Lo,
  Justin Johnson, and Georgia Gkioxari.
\newblock Accelerating 3d deep learning with pytorch3d.
\newblock \emph{arXiv preprint arXiv:2007.08501}, 2020.

\bibitem[Sadekar et~al.(2022)Sadekar, Tiwari, and Raman]{sadekar2022shadow}
Kaustubh Sadekar, Ashish Tiwari, and Shanmuganathan Raman.
\newblock Shadow art revisited: a differentiable rendering based approach.
\newblock In \emph{Proceedings of the IEEE/CVF Winter Conference on
  Applications of Computer Vision}, pages 29--37, 2022.

\bibitem[Sugihara(2018)]{sugihara2018topology}
Kokichi Sugihara.
\newblock Topology-disturbing objects: A new class of 3d optical illusion.
\newblock \emph{Journal of Mathematics and the Arts}, 12\penalty0 (1):\penalty0
  2--18, 2018.

\bibitem[Taubin(1995)]{taubin1995curve}
Gabriel Taubin.
\newblock Curve and surface smoothing without shrinkage.
\newblock In \emph{Proceedings of IEEE international conference on computer
  vision}, pages 852--857. IEEE, 1995.

\bibitem[Van~Geert and Wagemans(2024)]{van2024pragnanz}
Eline Van~Geert and Johan Wagemans.
\newblock Pr{\"a}gnanz in visual perception.
\newblock \emph{Psychonomic Bulletin \& Review}, 31\penalty0 (2):\penalty0
  541--567, 2024.

\bibitem[Zhang et~al.(2020)Zhang, Yin, Nie, and Zheng]{zhang2020deep}
Qing Zhang, Gelin Yin, Yongwei Nie, and Wei-Shi Zheng.
\newblock Deep camouflage images.
\newblock In \emph{Proceedings of the AAAI conference on artificial
  intelligence}, pages 12845--12852, 2020.

\end{thebibliography}
}


\twocolumn[{%
\begin{center}
{\Large\textbf{Supplementary Material for Mirror Illusion Art}}
\end{center}
\bigskip
}]

\renewcommand{\thesection}{S\arabic{section}}
\renewcommand{\theHsection}{S.\arabic{section}}
\renewcommand{\thefigure}{S\arabic{figure}}
\renewcommand{\thetable}{S\arabic{table}}
\setcounter{section}{0}
\setcounter{figure}{0}
\setcounter{table}{0}
\section{Supplementary Video}
See ``Demo Video.mp4'' for a full-angle visualization of mirror illusion 3D objects.

\section{From Voxel to Mesh} \label{sec: mesh}

Although voxel representation is convenient for 3D optimization, we convert the optimized 3D voxel model $V$ into a 3D mesh model to obtain a smoother representation suitable for 3D printing. To achieve this, we first use the classic Marching Cubes \cite{lorensen1998marching} algorithm, which extracts an isosurface from the voxel density field to generate the mesh. However, applying this method alone often leaves jagged artifacts on the mesh surface. To address this, we further apply the Taubin smoothing \cite{taubin1995curve} technique. For any vertex $y$ on the mesh, let $\bar{y}$ denote the mean position of all its neighboring vertices; then, the smoothed vertex coordinate $y'$ after Taubin smoothing is given by:
\begin{equation}
    y^{\prime}=(1-\xi)\cdot y+\xi\cdot\bar{y}.
\end{equation}
Here, $\xi$ is a hyperparameter that controls the degree of smoothing. Through this approach, we obtain a much smoother 3D mesh suitable for 3D printing.

\section{From Digital to Physical} \label{sec: phy}
To transform our design into a real-world artwork, we employ 3D printing technology to fabricate the optimized 3D mesh model.  We use the Mimaki 3DUJ 553 full-color 3D printer, which utilizes UV-cured inkjet printing technology. This device forms objects layer by layer by jetting micro-droplets of liquid photosensitive resin and instantly curing them with ultraviolet light, directly producing physical objects with textures and colorsm (no post-processing is required). The printing material is a water-soluble resin, which can be washed away after printing, facilitating the formation of complex structures. The layer thickness is set to approximately 20 $\mu$m, and the average model height is about 7 cm per piece. For each batch, we print five 3D models simultaneously, with an average printing time of about 11 hours.

\begin{figure*}[htbp]
\centering
\includegraphics[width=2\columnwidth]{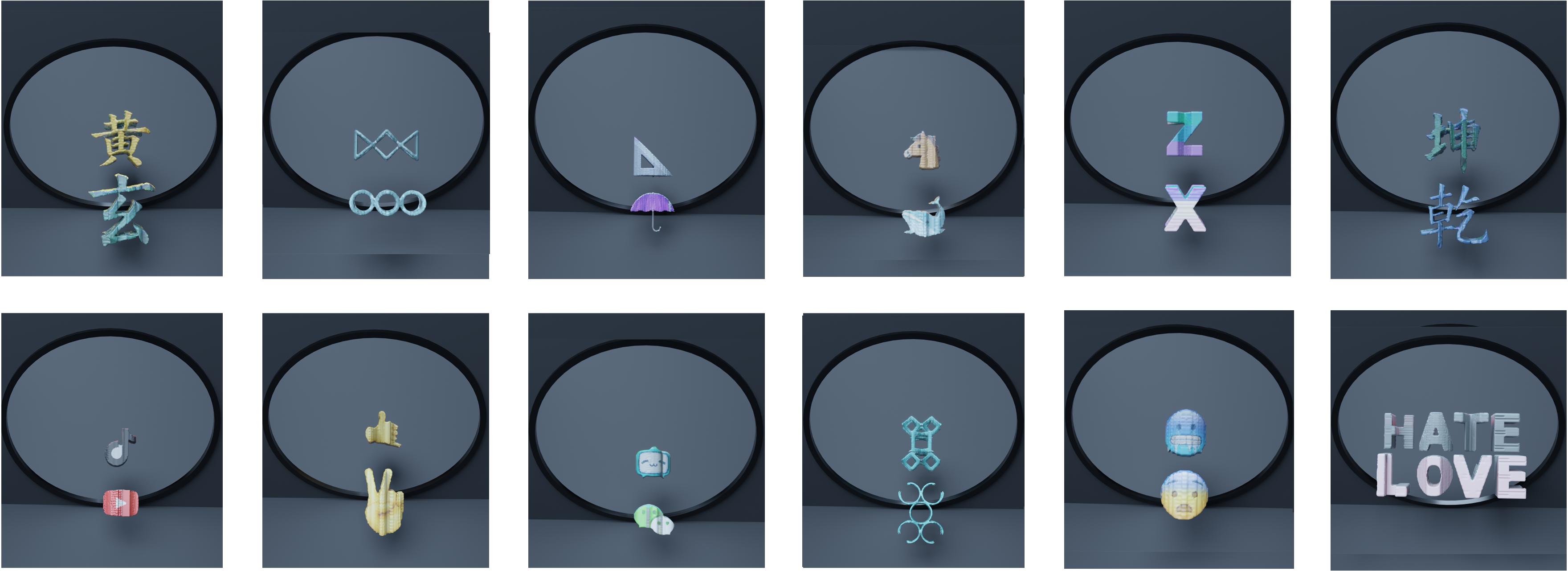} 
\caption{ More Digital visualization Examples of Mirror Illusion Arts designed by our AutoMIA method.}
\label{more_visual}
\end{figure*}

\section{Evaluation Metrics}

We employ the following evaluation metrics to systematically assess the 3D generation quality of Mirror Shadow Art, covering shape consistency, color consistency, surface noise intensity, and surface smoothness.

Shape Score (SS) is used to evaluate the shape consistency between the 3D object's projections and the supervision images. We denote $M^\mathrm{direct}$ and $M^\mathrm{mirror}$ as the projection masks of the 3D object in the direct and mirror directions, respectively, and $M_A$ and $M_B$ as the corresponding supervision pattern masks. The Shape Score is defined as follows:
\begin{equation}
    SS=\frac{\mathrm{IoU}(M^\mathrm{direct},M_A)+ \mathrm{IoU} (M^\mathrm{direct},M_A)}{2}.
\end{equation}

Color Score (CS) is used to evaluate the color consistency between the 3D object's projections and the supervision images. Based on Equation 3 of the main paper, we compute the average L1 loss between the projected color and the reference image color for the front view ($C_\mathrm{direct}$) and the mirrored view ($C_\mathrm{mirror}$), denoted as $L_{\text{color}}^\mathrm{direct}$ and $L_{\text{color}}^\mathrm{mirror}$, respectively. The Color Score can then be calculated as follows:
\begin{equation}
    CS=\frac{L_\mathrm{color}^\mathrm{direct}+L_\mathrm{color}^\mathrm{mirror}}{2}.
\end{equation}

Noise Level (NL) is used to evaluate the intensity of surface noise in the 3D voxel model. Let $\rho_{\text{origin}}$ denote the original density distribution of the 3D object. We apply a 3D Gaussian filter $G_\sigma$ with kernel radius $\sigma$ to obtain the smoothed density distribution $\rho_{\text{Gauss}} = G_\sigma(\rho_{\text{origin}})$. The surface noise distribution $R$ is then calculated as the difference between the original and smoothed densities, i.e., $R = \rho_{\text{origin}} - \rho_{\text{Gauss}}$. The Noise Level is computed as follows:
\begin{equation}
    NL=\frac{\|R\|_2^2}{\|\rho_{origin}\|_2^2}.
\end{equation}

Smooth Level (SL) is used to evaluate the surface smoothness of the 3D mesh model. Suppose the normal vectors of any two adjacent faces on the 3D mesh surface are $n_1$ and $n_2$, respectively, then the cosine of the angle between them is given by $n_1 \cdot n_2$. The Smooth Level can be calculated as follows:
\begin{equation}
    SL=\mathrm{Mean} (\frac{n_1\cdot n_2+1}{2}).
\end{equation}
Here, $\mathrm{Mean}$ denotes the average taken over all pairs of adjacent faces. 

SS ranges from $0$ to $1$, with higher values indicating greater shape similarity. CS ranges from $0$ to $1$, with lower values indicating greater color similarity. NL ranges from $0$ to $1$, with lower values reflecting less surface noise. SL ranges from $0$ to $1$, with higher values indicating a smoother mesh surface.

\section{Experimental Settings} \label{sm:setting}

To ensure fair comparison, we adopt the same experimental settings for all experiments. The resolution of the 3D voxel model is set to $128 \times 128 \times 128$. Rendering is performed using PyTorch3D \cite{ravi2020accelerating}, 
with the light source configured in direct illumination mode and the Grid Raysampler used as the ray sampler, where each ray samples 150 points. During rendering, the virtual viewing angle is randomly sampled between 22.5$^\circ$ and 60$^\circ$, which aligns with the real-world physical setup. The Adam \cite{kingma2014adam} optimizer is used for optimization, with 1000 epochs and a learning rate of 0.05. 

\section{Digital Visualization} \label{sm:digital visual}

We constructed a virtual environment for digital visualization of Mirror Illusion Art, based on Blender 4.5.  First, we created a virtual circular mirror with length, width, and height ratios of 12.5:12.5:1.3. The mirror material was set to a metallic value of 1, roughness of 0.04, refractive index of 1.5, and opacity of 1.0. Next, the 3D objects optimized by the AutoMIA method were imported into Blender and positioned in front of the virtual mirror. The virtual camera’s viewing angle was then adjusted to observe the Mirror Illusion effect, with specific requirements for camera angles discussed in Section 4.10. We used Blender's default renderer, setting the denoising filter threshold to 0.1 and the maximum number of samples to 2048.

Figure \ref{more_visual} provides more visualization examples based on this digital environment. All of the examples are generated by our AutoMIA method. 

\begin{table}[bp]
\centering
\small
\caption{The effect of volume size}\label{volume size}
\begin{tabular}{cccc}
\toprule[1.1pt]  
Volume Size & 64 & 128 & 256 \\
\hline
SL~$\uparrow$ & 0.979 & 0.989 & 0.977 \\
NL~$\downarrow$ & 0.059 & 0.050 & 0.021 \\
SS~$\uparrow$ & 0.929 & 0.931 & 0.972 \\
CS~$\uparrow$ & 0.019 & 0.018 & 0.018 \\
Time & 61s & 76s & 185s \\
Memory & 1978MB & 2672MB & 6952MB \\
\bottomrule[1.1pt] 
\end{tabular}
\end{table}

\begin{table*}[htbp]
\centering
\small
\caption{The effect of ray sampling density}\label{ray sampling}
\begin{tabular}{cccccc}
\toprule[1.1pt]  
Volume Size & 80 & 115 & 150 & 185 & 220 \\
\hline
SL~$\uparrow$ & 0.935 & 0.984  & 0.989  & 0.991  & 0.974 \\
NL~$\downarrow$ & 0.066 & 0.057  & 0.050  & 0.060  & 0.050 \\
SS~$\uparrow$ & 0.917 & 0.915  & 0.931  & 0.908  & 0.684 \\
CS~$\uparrow$ & 0.102 & 0.020  & 0.018  & 0.018  & 0.267  \\
Time & 71s   & 72s    & 76s    & 78s    & 80s \\
Memory & 2300MB  & 2520MB & 2672MB & 3036MB & 3904MB \\
\bottomrule[1.1pt] 
\end{tabular}
\end{table*}

\section{Underlying Mechanisms of Mirror Illusion Art}

Let’s delve into the underlying mechanisms from a neuroscience and psychology perspective. According to the Bayesian theory of perception \cite{knill1996perception}, the human visual system operates with prior experiences; in other words, our brain has an automatic completion mechanism. We first predict what we expect to see based on past experience and then update our perception using new sensory input. When a contour from a particular viewpoint resembles a familiar pattern, we tend to ``fill in'' the expected object in our mind.  The Biased Competition Principle \cite{beck2009top} suggests that our brain generates multiple possible interpretations for the same scene, while attention selectively enhances certain candidates, allowing them to dominate perception. Then what is the most likely interpretation? According to the Law of Prägnanz \cite{van2024pragnanz}, among all possible interpretations after this mental completion, our brain prefers the simplest, most symmetrical, and most stable patterns, similar to the principle of Occam’s Razor. This explains why we naturally expect the pattern in the mirror to be symmetrical and consistent with the one in front of the mirror. However, Mirror Illusion Art deliberately violates this intuitive expectation, creating a perceptual effect that surprises and challenges our usual ways of interpreting mirrored scenes.

\section{Volume Size}

During the optimization process, the volume size of the voxel model determines the level of detail in the reconstructed patterns, but also affects optimization time and memory usage. To quantitatively assess the impact of volume size on both reconstruction quality and computational cost, we tested three typical volume sizes: 64, 128, and 256. Following the experimental settings in Section \ref{sm:digital visual}, the results are shown in Table \ref{volume size}. The findings indicate that lower resolutions ($64 \times 64 \times 64$) reduce optimization time and memory consumption, but at the expense of reconstruction quality, while higher resolutions ($256 \times 256 \times 256$) offer only marginal improvements in quality with significantly increased computational costs. Therefore, in our experiments, the volume size is set to 128 to balance reconstruction quality and optimization efficiency.

\begin{figure}[tbp]
\centering
\includegraphics[width=1\columnwidth]{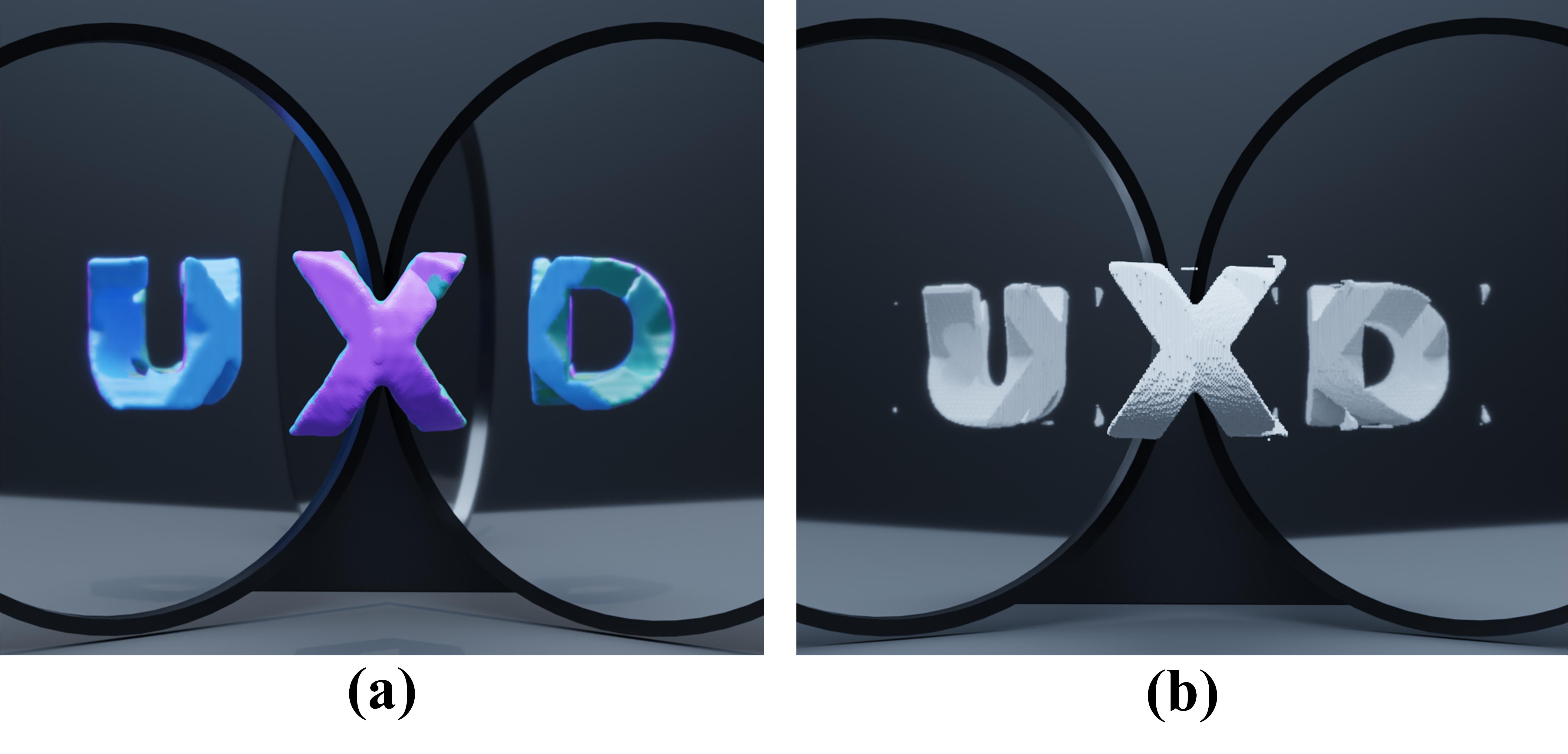} 
\caption{ 3-fold (non-orthogonal) mirror illusion art generated by (a) AutoMIA (Ours) and (b) Shadow Art (SA). Angle between two mirrors is 110$^\circ$.  Note that the SA-generated object has many artifacts (noise) on the surface and in the background. }
\label{3-fold}
\end{figure}

\section{Ray Sampling Density}

An important hyperparameter in the 3D rendering process is the ray sampling density, which directly affects both the quality of the rendered 3D object and the computational cost. To further quantify its impact, we selected five representative ray sampling densities: 80, 115, 150, 185, and 220. All other experimental settings followed Section \ref{sm:digital visual}. The results are shown in Table \ref{ray sampling}. Overall, increasing the sampling density improves the quality of the rendered 3D object, but also increases computation time and memory consumption. We also observed that excessively high sampling density can even reduce some reconstruction metrics, such as SL and SS. Therefore, to balance reconstruction quality and computational efficiency, we set the sampling density to 150 in our experiments.

\section{Three-Fold Mirror Illusion Art}
We further show that our method supports 3D reconstruction from three non-orthogonal views (3-fold images) (Figure \ref{3-fold}), producing richer colors, smoother surfaces, and less noise than Shadow Art—beneficial for high-fidelity reconstruction and physical printing. This further suggests that AutoMIA opens the door to a broader range of innovative applications beyond its current scope.

\end{document}